\documentclass[manuscript,screen]{acmart}

\usepackage{url}
\usepackage{threeparttable}

\setcopyright{none}
\begin{document}

\title{Infinite Video Understanding}

\author{Dell Zhang}
\orcid{0000-0002-8774-3725}
\affiliation{%
  \institution{Institute of Artificial Intelligence (TeleAI), China Telecom}
  \city{Shanghai}
  \country{China}
}
\email{dell.z@ieee.org}
\authornote{Joint first-authors with equal contributions.}

\author{Xiangyu Chen}
\orcid{0000-0003-2156-4959}
\affiliation{%
  \institution{Institute of Artificial Intelligence (TeleAI), China Telecom}
  \city{Shanghai}
  \country{China}
}
\email{chenxy178@chinatelecom.cn}
\authornotemark[1]

\author{Jixiang Luo}
\orcid{0009-0002-4698-7835}
\affiliation{%
  \institution{Institute of Artificial Intelligence (TeleAI), China Telecom}
  \city{Shanghai}
  \country{China}
}
\email{luojx14@chinatelecom.cn}

\author{Mengxi Jia}
\orcid{0000-0002-0979-9803}
\affiliation{%
  \institution{Institute of Artificial Intelligence (TeleAI), China Telecom}
  \city{Shanghai}
  \country{China}
}
\email{jiamx1@chinatelecom.cn}

\author{Changzhi Sun}
\orcid{0009-0003-3123-3499}
\affiliation{%
  \institution{Institute of Artificial Intelligence (TeleAI), China Telecom}
  \city{Shanghai}
  \country{China}
}
\email{czsun@chinatelecom.cn}

\author{Ruilong Ren}
\orcid{0009-0007-2017-2159}
\affiliation{%
  \institution{Institute of Artificial Intelligence (TeleAI), China Telecom}
  \city{Shanghai}
  \country{China}
}
\affiliation{%
  \institution{Peking University}
  \city{Beijing}
  \country{China}
}
\email{ruilongren@163.com}
\authornote{Work done during an internship with TeleAI.}

\author{Jingren Liu}
\orcid{0009-0009-0163-4105}
\affiliation{%
  \institution{Institute of Artificial Intelligence (TeleAI), China Telecom}
  \city{Shanghai}
  \country{China}
}
\affiliation{%
  \institution{Tianjin University}
  \city{Tianjin}
  \country{China}
}
\email{chengxuyuangg@163.com}
\authornotemark[2]

\author{Hao Sun}
\orcid{0009-0007-7917-1628}
\affiliation{%
  \institution{Institute of Artificial Intelligence (TeleAI), China Telecom}
  \city{Shanghai}
  \country{China}
}
\email{sunh10@chinatelecom.cn}

\author{Xuelong Li}
\orcid{0000-0002-0019-4197}
\affiliation{%
  \institution{Institute of Artificial Intelligence (TeleAI), China Telecom}
  \city{Shanghai}
  \country{China}
}
\email{xuelong_li@ieee.org}
\authornote{Corresponding author.}

\renewcommand{\shortauthors}{Zhang et al.}

\begin{abstract}
The rapid advancements in Large Language Models (LLMs) and their multimodal extensions (MLLMs) have ushered in remarkable progress in video understanding. 
However, a fundamental challenge persists: effectively processing and comprehending video content that extends beyond minutes or hours. 
While recent efforts like Video-XL-2 have demonstrated novel architectural solutions for extreme efficiency, and advancements in positional encoding such as HoPE and VideoRoPE++ aim to improve spatio-temporal understanding over extensive contexts, current state-of-the-art models still encounter significant computational and memory constraints when faced with the sheer volume of visual tokens from lengthy sequences. 
Furthermore, maintaining temporal coherence, tracking complex events, and preserving fine-grained details over extended periods remain formidable hurdles, despite progress in agentic reasoning systems like Deep Video Discovery.
This position paper posits that a logical, albeit ambitious, next frontier for multimedia research is Infinite Video Understanding --- the capability for models to continuously process, understand, and reason about video data of arbitrary, potentially never-ending duration. 
We argue that framing Infinite Video Understanding as a blue-sky research objective provides a vital north star for the multimedia, and the wider AI, research communities, driving innovation in areas such as streaming architectures, persistent memory mechanisms, hierarchical and adaptive representations, event-centric reasoning, and novel evaluation paradigms. 
Drawing inspiration from recent work on long/ultra-long video understanding and several closely related fields, we outline the core challenges and key research directions towards achieving this transformative capability.
\end{abstract}

\maketitle

\section{Introduction}

Multimedia research stands at the intersection of diverse digital modalities (including images, text, video, audio, speech, music, and sensor data). 
Video, in particular, constitutes a vast and ever-growing repository of information, with online platforms like YouTube and TikTok receiving uploads measured in thousands per second. 
Understanding this type of content, inferring semantic meaning, and enabling intelligent interaction have been central goals of computer science~\cite{narasimhanMultimodalLongTermVideo2023}.

The advent of Large Language Models (LLMs) and their integration with visual encoders to form Multimodal Large Language Models (MLLMs) have significantly advanced our capabilities in areas such as video captioning and question-answering~\cite{wengLongVLMEfficientLong2024,qianStreamingLongVideo2024,zouSecondsHoursReviewing2024,shenLongVUSpatiotemporalAdaptive2024}. 
These models leverage the powerful textual generation and reasoning abilities of LLMs to process and interpret visual information.

However, the vast majority of current research and benchmarks for video understanding, even those labeled ``long-form'', typically focus on videos measured in minutes or, at most, a few hours~\cite{mangalamEgoSchemaDiagnosticBenchmark2023,liMVBenchComprehensiveMultimodal2024,fuVideoMMEFirstEverComprehensive2024,zhaoNeedleVideoHaystack2025,zhouMLVUBenchmarkingMultitask2025,wangLVBenchExtremeLong2024,wuLongVideoBenchBenchmarkLongcontext2024,chandrasegaranHourVideo1HourVideoLanguage2024,wuLongViTUInstructionTuning2025,yangEgoLifeEgocentricLife2025} (see Table~\ref{tab:longform_benchmark_summary}). 
While in recent years, significant progress has been made in handling videos of such lengths --- 
notably through advancements like Video-XL-2's task-aware KV sparsification for efficiency~\cite{qinVideoXL2VeryLongVideo2025} and agentic search strategies in Deep Video Discovery for complex reasoning over segmented clips~\cite{zhangDeepVideoDiscovery2025} --- 
the fundamental constraint of limited context windows inherent to many LLMs and the associated computational burden persist. 
Moreover, the intricate spatial-temporal structure of video demands specialized positional encoding, an area where HoPE~\cite{liHoPEHybridPosition2025} and VideoRoPE++~\cite{weiVideoRoPEWhatMakes2025} have recently demonstrated improved length generalization capabilities through novel frequency allocation and dynamic scaling. 
Processing long videos necessitates a substantial number of tokens, which dramatically increases computational demands and risks losing early contextual information~\cite{wengLongVLMEfficientLong2024,qianStreamingLongVideo2024,shuVideoXLExtraLongVision2024,shenLongVUSpatiotemporalAdaptive2024,chengScalingVideoLanguageModels2025}.


\begin{table*}[tb]
  \centering
  \begin{threeparttable}
  \caption{A summary of existing long video understanding benchmarks\tnote{1}. Most so-called ``long-form'' video datasets focus on videos of limited length (typically minutes to a few hours). }
  \begin{tabular}{@{}lcrrrrr@{}}
    \toprule
    \textbf{Dataset} & \textbf{Release Date} & \textbf{\#Clips} & \textbf{Avg. Length} & \textbf{Max. Length} & \textbf{Total Length} & \textbf{\#QA-Pairs} \\
    \midrule
    \textbf{MovieChat}\cite{songMovieChatDenseToken2024}                    & 2023-07 &     1,000 &      9.4 min &        21 min &        156 hours & 13,000 \\
    \textbf{EgoSchema}~\cite{mangalamEgoSchemaDiagnosticBenchmark2023}      & 2023-08 &     5,031 &        3 min &       3 min   &     251.24 hours &  5,031 \\
    \textbf{MVBench}~\cite{liMVBenchComprehensiveMultimodal2024}            & 2023-11 &     3,655 &    16.7 sec  &      1.93 min &      15.46 hours &  4,000 \\
    \textbf{VideoMME}~\cite{fuVideoMMEFirstEverComprehensive2024}           & 2024-05 &       900 &       17 min &        1 hour &     255.32 hours &   2,700 \\
    \textbf{VNBench-Long}~\cite{zhaoNeedleVideoHaystack2025}                & 2024-06 &     1,350 &    55.65 sec &      2.93 min &      20.87 hours &   5,400 \\    
    \textbf{MLVU}~\cite{zhouMLVUBenchmarkingMultitask2025}                  & 2024-06 &     1,337 &     14.5 min &       9 hours &     323.37 hours &   3,102 \\
    \textbf{LVBench}~\cite{wangLVBenchExtremeLong2024}                      & 2024-06 &       103 &   1.12 hours &    2.33 hours &     115.53 hours &   1,549 \\
    \textbf{LongVideoBench}~\cite{wuLongVideoBenchBenchmarkLongcontext2024} & 2024-07 &     3,761 &    12.18 min &     24.22 min &     763.69 hours &   6,678 \\
    \textbf{HourVideo}~\cite{chandrasegaranHourVideo1HourVideoLanguage2024} & 2024-11 &       500 &    45.73 min &    1.93 hours &     381.08 hours &  12,976 \\
    \textbf{LongViTU}~\cite{wuLongViTUInstructionTuning2025}                & 2025-01 &     1,833 &     29.3 min &    2.01 hours &        900 hours & 121,143 \\
    \textbf{EgoLifeQA\tnote{2}}\ \ \cite{yangEgoLifeEgocentricLife2025}     & 2025-03 & 6 (3,000) &   44.3 hours &    55.2 hours &        266 hours &   6,000 \\
    \bottomrule
  \end{tabular}
  \label{tab:longform_benchmark_summary}
\begin{tablenotes}

{\footnotesize
\item [1] Statistics are calculated based on the actual downloaded video data. 
\item [2] EgoLifeQA's 6 long videos are organized into 3,000 video clips, with most QA pairs based on these short clips.
}
\end{tablenotes}
\end{threeparttable}
\end{table*}

We propose that the ultimate challenge and a critical blue-sky research topic for the multimedia community is to achieve \emph{Infinite Video Understanding}. 
This is not merely an incremental improvement in handling longer videos. 
It is a conceptual shift towards building models capable of processing and understanding video streams of arbitrary, potentially unbounded duration. 
This includes processing live video feeds, continuously generated video content, or simply video archives so vast that they approximate infinity from a processing perspective. 
Drawing inspiration from recent discussions regarding understanding long/ultra-long videos, we envision intelligent systems that can build and maintain a coherent understanding of video content over hours, days, weeks, months, years, or even indefinitely, continuously updating their knowledge and reasoning capabilities.

Framing Infinite Video Understanding as a research \emph{north star} provides a compelling, long-term goal that necessitates fundamentally new approaches beyond merely scaling up existing techniques. 
This position paper will elaborate on the profound challenges inherent in current long video understanding and articulate the key research directions required to embark on the journey towards Infinite Video Understanding.

\section{The Enduring Challenges of Long Video Understanding}

Before charting a course towards understanding infinitely long videos, it is crucial to fully appreciate the significant hurdles that limit current models even on videos of finite, albeit long, duration. 
These challenges stem primarily from the discrepancy between the continuous, high-density nature of video data and the discrete, context-limited processing capabilities of current sequence models including MLLMs.
\begin{itemize}
    \item \textbf{Context Window Limitations}. 
        LLMs/MLLMs, while powerful, operate with finite context windows, typically ranging from 4K to 128K tokens~\cite{chengScalingVideoLanguageModels2025}, although some models are pushing towards 1M or even 10M tokens~\cite{teamGemini15Unlocking2024,liuWorldModelMillionLength2025,liuVideoXLProReconstructiveToken2025}. 
        Video data, at standard frame rates (e.g., 24 FPS), would generate an enormous number of visual tokens.
        For example, a single minute of video can easily produce over a million tokens. 
        This rapidly exceeds the effective context capacity of LLMs/MLLMs, making it difficult to maintain a global understanding or refer to distant events within the video~\cite{qianStreamingLongVideo2024,shuVideoXLExtraLongVision2024,chengScalingVideoLanguageModels2025}.
        While methods like VideoRoPE++~\cite{weiVideoRoPEWhatMakes2025} have introduced 3D position embedding structures with adjustable temporal spacing to better preserve spatio-temporal relationships over long contexts, and HoPE~\cite{liHoPEHybridPosition2025} proposes a hybrid frequency allocation and dynamic temporal scaling to improve length generalization, the inherent challenge of token explosion remains. 
        These efforts contribute to making the context windows more effective rather than simply larger.
    \item \textbf{Memory Burdens and Computational Costs}. 
        Representing long videos with a large number of tokens leads to significant memory footprints and prohibitive computational costs. 
        As video length increases, traditional token-based approaches become increasingly infeasible, necessitating the development of more efficient representations and scalable architectures~\cite{wengLongVLMEfficientLong2024,qianStreamingLongVideo2024,shuVideoXLExtraLongVision2024,shenLongVUSpatiotemporalAdaptive2024,chengScalingVideoLanguageModels2025}. 
        Recent work, exemplified by Video-XL-2~\cite{qinVideoXL2VeryLongVideo2025}, directly tackles this by proposing task-aware KV sparsification through chunk-based pre-filling and bi-level KV decoding, enabling processing of extremely long sequences (e.g., up to 10,000 frames) on single GPUs with high efficiency.
    \item \textbf{Information Loss from Downsampling or Compression}. 
        To mitigate the token explosion, many existing approaches rely on sparse temporal sampling or aggressive frame compression. 
        However, these strategies often come at the cost of discarding fine-grained spatial or contextual details, undermining the model's understanding of the video content~\cite{wengLongVLMEfficientLong2024,qianStreamingLongVideo2024,chengScalingVideoLanguageModels2025}. 
        While hierarchical compression and adaptive strategies aim to preserve more relevant visual information, balancing compression ratio and information fidelity remains a difficult problem.
        While hierarchical compression and adaptive strategies aim to preserve more relevant information, such as Video-XL-2's Dynamic Token Synthesize module for spatio-temporal redundancy compression or its bi-level KV decoding that selectively loads dense KVs for relevant chunks~\cite{qinVideoXL2VeryLongVideo2025}, balancing compression ratio and information fidelity remains a difficult challenge~\cite{qianStreamingLongVideo2024,shuVideoXLExtraLongVision2024,shenLongVUSpatiotemporalAdaptive2024,liVideoChatFlashHierarchicalCompression2025,chengScalingVideoLanguageModels2025,weiVideoRoPEWhatMakes2025}. 
        Furthermore, approaches like Deep Video Discovery~\cite{zhangDeepVideoDiscovery2025} mitigate information loss by retaining original decoded frames alongside text captions, enabling detailed pixel-level inspection when required by the agent.
    \item \textbf{Maintaining Temporal Coherence and Tracking Long-Term Dependencies}. 
        Videos of substantial length consist of multiple events and complex activities with dependencies that may span long time intervals. 
        Current models struggle to maintain temporal coherence and track the story-line or evolution of events across sequential segments~\cite{zouSecondsHoursReviewing2024,wengLongVLMEfficientLong2024,qianStreamingLongVideo2024}. 
        Significant work is ongoing to address this, with HoPE~\cite{liHoPEHybridPosition2025} introducing dynamic temporal scaling to learn multi-scale temporal relationships for robust modeling, and VideoRoPE++~\cite{weiVideoRoPEWhatMakes2025} emphasizing low-frequency temporal allocation to mitigate periodic oscillations and ensure temporal consistency over extended contexts. 
        Additionally, Deep Video Discovery~\cite{zhangDeepVideoDiscovery2025} focuses on event-centric understanding through agentic reasoning, enabling the model to track and resolve entities and actions across a video's extensive duration, even when separated by large temporal gaps.
    \item \textbf{Handling Noisy and Misaligned Data}. 
        Real-world long videos, particularly instructional or user-generated content, often contain noisy automatically generated transcripts with repetitive phrases and misalignments between visual and auditory information. 
        For example, demonstrators might describe actions before or after performing them, complicating the alignment of modalities~\cite{narasimhanMultimodalLongTermVideo2023}. 
        The Lengthy Multimodal Stack subtask in the V-RULER benchmark introduced by VideoRoPE++~\cite{weiVideoRoPEWhatMakes2025} explicitly challenges models to isolate relevant visual information amidst extensive irrelevant textual noise, highlighting the importance of robust multimodal processing in such conditions. 
        Addressing these data quality issues is crucial for robust understanding of long videos.
    \item \textbf{Evaluating Understanding at Scale}. 
        Developing reliable benchmark datasets and evaluation metrics for understanding video content that spans hours or even approaches infinite length is a significant challenge. 
        Recent progress includes the V-RULER benchmark introduced by VideoRoPE++~\cite{weiVideoRoPEWhatMakes2025,hsiehRULERWhatsReal2024}, which features challenging subtasks like Needle Retrieval under Distractors (NRD) and Multi-Key, Multi-Value (MKMV) to assess fine-grained temporal localization, entity tracking, and robustness against distractors in long videos. 
        Deep Video Discovery~\cite{zhangDeepVideoDiscovery2025} demonstrates state-of-the-art performance on LVBench~\cite{wangLVBenchExtremeLong2024} and evaluates across LongVideoBench~\cite{wuLongVideoBenchBenchmarkLongcontext2024}, Video-MME~\cite{fuVideoMMEFirstEverComprehensive2024}, and EgoSchema~\cite{mangalamEgoSchemaDiagnosticBenchmark2023}. 
        Similarly, HoPE!\cite{liHoPEHybridPosition2025} and Video-XL-2~\cite{qinVideoXL2VeryLongVideo2025} also provide comprehensive evaluations on LongVideoBench~\cite{wuLongVideoBenchBenchmarkLongcontext2024}, Video-MME~\cite{fuVideoMMEFirstEverComprehensive2024} and MLVU~\cite{zhouMLVUBenchmarkingMultitask2025} across extended context lengths. 
        These benchmarks, while growing in length, still do not fully capture the complexities of continuous, potentially never-ending streams or tasks requiring reasoning over truly vast temporal horizons~\cite{zouSecondsHoursReviewing2024,chenLongVILAScalingLongContext2024,chengScalingVideoLanguageModels2025,wuLongViTUInstructionTuning2025,luoVideoRAGVisuallyalignedRetrievalAugmented2024,liuVideoXLProReconstructiveToken2025,chandrasegaranHourVideo1HourVideoLanguage2024,narasimhanMultimodalLongTermVideo2023}. 
        Evaluating fine-grained details or comprehensive contextual understanding over such scales requires further novel approaches.
\end{itemize}

These challenges highlight that while ``long video understanding'' (on the scale of minutes/hours) is an active and important area of research, it is still grappling with fundamental limitations that prevent scaling to truly arbitrary or infinite lengths.

\section{The Vision: Towards Infinite Video Understanding}

The pursuit of Infinite Video Understanding signifies a fundamental departure from traditional, bounded paradigms of video comprehension, aiming instead for continuous, streaming-level semantic interpretation across arbitrary temporal scales. 
Realizing such an ambitious objective necessitates not only scalable encoder architectures but also robust theoretical frameworks to inform model design. 
These frameworks must accommodate incremental updates, abstracted memory management, long-range information retrieval, and semantic adaptability~\cite{zouSecondsHoursReviewing2024}.

Let the video stream be represented as a continuous-time function \( V : \mathbb{R}_{\geq 0} \rightarrow \mathcal{X} \), where $\mathbb{R}{\geq 0}$ denotes the non-negative real line (representing time), and $\mathcal{X} \subseteq \mathbb{R}^{H \times W \times C}$ represents the space of visual frames with spatial resolution $H \times W$ and $C$ color channels.

An ideal encoder $\mathcal{E}$ processes each incoming frame as follows: 
\begin{equation} 
    z_t = \mathcal{E}(V(t), \mathcal{M}{<t}, \theta), 
\end{equation}
where $z_t \in \mathcal{Z}$ denotes the latent representation at time $t$, $\mathcal{M}{<t}$ denotes the memory state accumulated up to but not including time $t$, and $\theta$ represents the encoder’s learnable parameters.

To be compatible with infinite-length streaming, the encoder must satisfy three operational constraints:
\begin{itemize}
    \item \textbf{Incremental Encoding}: \( z_t \) must be computable without accessing \( V(\tau) \) for \( \tau > t \)~\cite{qianStreamingLongVideo2024}.
    \item \textbf{Memory Consolidation}: \( \mathcal{M}_{<t} \gets f(\mathcal{M}_{<t-1}, z_{t-1}) \), maintaining bounded memory~\cite{santos$infty$VideoTrainingFreeApproach2025}.
    \item \textbf{Query-Aligned Retrieval}: Efficient access to relevant \( z_{\tau} \) for \( \tau \ll t \) given a query context \( q_t \)~\cite{luoVideoRAGVisuallyalignedRetrievalAugmented2024}.
\end{itemize}

To tackle these, visual encoder architectures have evolved into three prominent design paradigms: 
(1) self-supervised video pretraining, 
(2) visual tokenization, and 
(3) vision-language alignment. 
Each paradigm offers distinct inductive biases and computational trade-offs relevant to persistent video comprehension.

\textbf{Self-Supervised Pretraining on Raw Videos:}
Self-supervised video representation learning leverages temporal consistency and reconstruction objectives to learn rich spatiotemporal features without explicit labels. 
Formally, for a video stream \( V: \mathbb{R}_{\geq 0} \), an encoder \( \mathcal{E} \) learns latent representations \( z_t \in \mathcal{Z} \) such that:
\begin{equation}
    z_t = \mathcal{E}(V(t), \mathcal{M}_{<t}),
\end{equation}
where \( \mathcal{M}_{<t} \) denotes abstracted memory until time \( t \).

A typical learning objective in masked autoencoding frameworks such as VideoMAE~\cite{tongVideoMAEMaskedAutoencoders2022}, VideoMAE V2~\cite{wangVideoMAEV2Scaling2023}, CropMAE~\cite{eymaelEfficientImagePreTraining2024}, SiamMAE~\cite{guptaSiameseMaskedAutoencoders2023} or CrossVideoMAE~\cite{ahamedCrossVideoMAESelfSupervisedImageVideo2025} is:
\begin{equation}
    \mathcal{L}_{\text{MAE}} = \mathbb{E}_{x_t \sim V} \left\| x_t^{\text{masked}} - \mathcal{D}(\mathcal{E}(x_t^{\text{visible}})) \right\|^2,
\end{equation}
where $x_t^{\text{visible}}$ and $x_t^{\text{masked}}$ refer to the observed and occluded regions of frame $x_t$, and $\mathcal{D}$ is a decoder reconstructing the masked content from latent embeddings.

Several architectures extend this approach by modeling motion-structure disentanglement. For instance, VidTwin~\cite{wangVidTwinVideoVAE2025} decomposes the latent space as:
\begin{equation} 
z_t = \left[ z_t^{\text{struct}}; z_t^{\text{dyn}} \right], \quad \text{with } z_t^{\text{struct}} \perp z_t^{\text{dyn}}, 
\end{equation} 
where $z_t^{\text{struct}}$ encodes static scene structure, $z_t^{\text{dyn}}$ captures dynamic motion, and $\perp$ denotes a design constraint enforcing statistical orthogonality.

Moreover, stochastic modeling techniques have been introduced to anticipate multi-modal futures by modeling latent uncertainty in forthcoming frames~\cite{jangVisualRepresentationLearning2024}.

\textit{Strengths.} 
These models are capable of learning semantically rich and temporally coherent representations directly from raw video data, enabling scalable training over massive unlabeled corpora.

\textit{Limitations.} 
Despite their efficiency, such models often lack high-level semantic abstraction and consistent memory representations over long temporal horizons. 
Furthermore, in practice, most frameworks necessitate keyframe sampling to mitigate computational burden and stabilize optimization, which may result in underrepresentation of transient or contextually critical events.

\textbf{Visual Tokenization and Discrete Representation Learning:} 
While no dedicated tokenizer has yet been established specifically for scalable, general-purpose video understanding, several methods have explored visual tokenization within the domain of video generation, demonstrating promising capabilities in compressing and abstracting temporal visual information.
\begin{equation} 
z_t = \mathcal{T}_v(V_t), \quad z_t \in \mathbb{Z}^d,
\end{equation} 
where $\mathcal{T}_v$ denotes a visual tokenizer, and $\mathbb{Z}^d$ is a discrete latent space.

Models such as MAGVIT~\cite{yuMAGVITMaskedGenerative2023}, MAGVIT-v2~\cite{yuLanguageModelBeats2024}, and Cosmos~\cite{nvidiaCosmosWorldFoundation2025} demonstrate strong representational efficiency through discrete encodings. 
Signal-based alternatives like SNeRV~\cite{kimSNeRVSpectrapreservingNeural2025} leverage spectral representations to encode videos without explicit frame decoding, preserving temporal coherence in a continuous format.

\textit{Strengths.} 
Discrete tokens are inherently compact, compressible, and highly suitable for integration with language-based models. 
A critical advantage is their ability to operate without keyframe sampling, thereby providing temporally uniform abstraction over arbitrarily long sequences and enabling effective indexing and symbolic reasoning.

\textit{Limitations.} 
Most existing tokenizers are developed for generative tasks and thus lack fine-grained temporal resolution or semantic precision. 
Phenomena such as token drift and gradual information loss over extended horizons remain significant barriers to their deployment in continuous understanding settings.

\textbf{Vision-Language Pretraining and Alignment:} 
Vision-language pretrained models bridge visual inputs and text embeddings via shared latent spaces, enabling semantic alignment across modalities. 
Formally, a video frame \( V_t \) is encoded as:
\begin{equation} 
    z_t = \mathcal{E}(V_t), \quad \text{aligned via } \langle z_t, \mathcal{T}_l(w) \rangle \rightarrow \text{semantic matching},
\end{equation} 
where $\mathcal{T}_l(w)$ is the embedding of a text token $w$, and $\langle \cdot, \cdot \rangle$ denotes a similarity measure (e.g., dot product or cosine similarity).

Early large-scale models like CLIP~\cite{radfordLearningTransferableVisual2021} and SigLIP~\cite{zhaiSigmoidLossLanguage2023,tschannenSigLIP2Multilingual2025} have demonstrated the effectiveness of contrastive alignment between visual and linguistic modalities on image-level tasks. 
These methods provide a foundation for video-based extensions by promoting generalizable, zero-shot transferable representations.
More recent works such as VideoPrism~\cite{zhaoVideoPrismFoundationalVisual2025} extend this idea to continuous video inputs, aligning unified transformer-based encoders with frozen or learned text features. 
VideoTree~\cite{wangVideoTreeAdaptiveTreebased2025} introduces a hierarchical token structure that facilitates long-video summarization and efficient querying by large language models.

\textit{Strengths.} 
By projecting video features into a language-aligned embedding space, these models support zero-shot generalization, instruction-following, and semantically-aware retrieval—capabilities foundational to interactive and long-term video comprehension.

\textit{Limitations.} 
These models are typically trained offline using temporally coarse granularity (e.g., fixed-length clips or keyframes), limiting their temporal expressiveness. 
Moreover, the training pipeline often relies on keyframe selection to achieve coverage efficiency, potentially omitting fine-grained temporal cues. 
The reliance on large-scale paired data and high compute further impedes scalability for lifelong video modeling.

\section{Key Research Directions for Infinite Video Understanding}

Achieving Infinite Video Understanding requires significant fundamental research breakthroughs across multiple areas of computer science. 
We propose the following key research directions, building upon but extrapolating significantly from current work on long video understanding.
\begin{itemize}
    \item \textbf{Architectures for Persistent Memory and Knowledge Consolidation}. 
        Current approaches for long video understanding include segmenting videos and processing sequentially with memory propagation~\cite{qianStreamingLongVideo2024}, or using sparse memory mechanisms~\cite{santos$infty$VideoTrainingFreeApproach2025}. 
        For infinite video streams, models need perpetual memory systems. 
        This could involve continuously consolidating short-term observations into a more abstract, lower-dimensional long-term memory (LTM)~\cite{santos$infty$VideoTrainingFreeApproach2025}. 
        This LTM would need to be continuously updated and be queryable\cite{qianStreamingLongVideo2024}. 
        Concepts like Retrieval-Augmented Generation (RAG), already explored for long video understanding by using auxiliary texts derived from video content~\cite{luoVideoRAGVisuallyalignedRetrievalAugmented2024}, could be extended to retrieve relevant information from a vast, dynamically updated LTM store. 
        Deep Video Discovery~\cite{zhangDeepVideoDiscovery2025} offers an example of such a system, employing a multi-granular video database (global, clip, and frame levels) and an agentic search paradigm to query and retrieve information efficiently. 
        Similarly, Video-XL-2's bi-level KV decoding strategy~\cite{qinVideoXL2VeryLongVideo2025}, by selectively loading dense or sparse KVs based on task relevance, represents a form of intelligent memory management and information consolidation. 
        Research into neuro-inspired memory consolidation processes also offers a promising avenue~\cite{santos$infty$VideoTrainingFreeApproach2025}.
    \item \textbf{Streaming and Incremental Processing Paradigms}. 
        Standard batch processing is incompatible with infinite video streams, therefore architectures must be designed for efficient streaming and incremental processing. 
        This involves processing video chunks or frames as they arrive, updating the internal state or memory, and discarding raw data that is no longer needed but has been summarized or consolidated~\cite{qianStreamingLongVideo2024}. 
        Video-XL-2's chunk-based pre-filling~\cite{qinVideoXL2VeryLongVideo2025}, which processes visual tokens in equal-length chunks with full attention within and sparse attention across chunks, provides a highly efficient and scalable model for incremental processing. 
        Research is needed into how to effectively handle the transitions between video chunks and ensure global coherence is maintained without processing overlapping video segments extensively.
    \item \textbf{Hierarchical and Adaptive Spatiotemporal Representations}. 
        Managing the immense video data volume requires multi-resolution representations that adapt based on importance or context~\cite{qianStreamingLongVideo2024,shuVideoXLExtraLongVision2024,shenLongVUSpatiotemporalAdaptive2024,chengScalingVideoLanguageModels2025}. 
        Hierarchical compression approaches that represent information at different granularities (frame, clip, event, long-term summary) are a starting point~\cite{liVideoChatFlashHierarchicalCompression2025,chengScalingVideoLanguageModels2025}. 
        Deep Video Discovery~\cite{zhangDeepVideoDiscovery2025} exemplifies this with its multi-granular video database, encompassing global, clip, and frame-level information, enabling different search tools to operate at varying granularities. 
        For infinite video streams, these representations must be dynamically managed, perhaps increasing granularity for salient events and aggressively summarizing redundant periods~\cite{qianStreamingLongVideo2024,shuVideoXLExtraLongVision2024,shenLongVUSpatiotemporalAdaptive2024}. 
        Novel network architectures that explicitly encode hierarchical layering of information, such as atomic actions, steps, and larger events over indefinite durations, are crucial~\cite{narasimhanMultimodalLongTermVideo2023}. 
        Recent work on positional encoding, such as VideoRoPE++'s 3D structure and low-frequency temporal allocation~\cite{weiVideoRoPEWhatMakes2025} and HoPE's hybrid frequency allocation and dynamic temporal scaling,~\cite{liHoPEHybridPosition2025} directly contribute to learning more robust and flexible spatio-temporal representations over extended periods. 
        Task-specific abstraction levels should be integrated, tailoring representations for fine-grained actions, intermediate steps, or high-level narrative summaries.
    \item \textbf{Event-Centric Understanding and Tracking}. 
        From users' perspective, long videos are best understood as sequences of events, activities, and states that unfold over time~\cite{zouSecondsHoursReviewing2024,chenLongVILAScalingLongContext2024}. 
        Hence the models should prioritize robust event detection, classification, tracking, and understanding of causal and temporal relationships between events, even those separated by vast time spans. 
        VideoRoPE++'s Multi-Key Multi-Value (MKMV) task within its V-RULER benchmark~\cite{weiVideoRoPEWhatMakes2025,hsiehRULERWhatsReal2024} specifically addresses the need to track entity identity and interpret varying behavior across extended temporal intervals, requiring resolution of multiple semantic targets for a single entity in a long video stream. 
        Building a dynamic graph or knowledge representation of events and their attributes might be able to provide a structured way to manage video understanding over infinite time.
    \item \textbf{Handling Input Variability, Noise, and Concept Drift}. 
        Over infinitely long video streams, data quality may vary, sensors might fail temporarily, and the nature of the content or the underlying patterns might slowly change (concept drift)~\cite{narasimhanMultimodalLongTermVideo2023}. 
        Infinite Video Understanding models should be resilient to noise and capable of detecting and adapting to shifts in data distribution or content characteristics. 
        The Needle Retrieval under Distractors (NRD) subtask and the Lengthy Multimodal Stack subtask of VideoRoPE++'s V-RULER benchmark~\cite{weiVideoRoPEWhatMakes2025,hsiehRULERWhatsReal2024} explicitly evaluate model robustness against visually similar distractors and unrelated textual noise, respectively. 
        Research into robust online learning and adaptation methods could be essential. 
        Furthermore, approaches to clean noisy transcripts and improve alignment between modalities in real-time are needed~\cite{narasimhanMultimodalLongTermVideo2023}.
    \item \textbf{Integrated Multimodal Processing for Continuous Streams}. 
        Video streams often contain rich audio (speech) and textual information (overlaid text)~\cite{zouSecondsHoursReviewing2024}. 
        Thus, Infinite Video Understanding should seamlessly integrate these modalities into the continuous processing framework~\cite{narasimhanMultimodalLongTermVideo2023}. 
        This involves not just parallel processing but leveraging cross-modal dependencies and signals~\cite{zhouAVHashJointAudioVisual2024} to enhance understanding and robustly track information over time, especially when one modality might be temporarily degraded.
    \item \textbf{Deep Reasoning over Evolving Knowledge Augmented with Tools}. 
        As systems acquire and consolidate an ever-growing base of long-term knowledge, supporting deep reasoning capabilities becomes increasingly challenging. 
        Models must go beyond merely ``seeing'' and ``remembering'' --- they should be capable of reflection and prediction~\cite{chengVideoHolmesCanMLLM2025,fengVideoR1ReinforcingVideo2025}. 
        This entails developing mechanisms for both reflective reasoning over past events and predictive reasoning based on accumulated memory and contextual understanding. 
        Future systems should leverage continuously evolving internal representations to answer questions such as ``Why did this event happen?'' or ``What is likely to happen next, and why?''. 
        Such reasoning marks a shift in video understanding from the perceptual level to the cognitive level. 
        Furthermore, recent advances in multimodal reasoning (e.g., OpenAI's GPT-4o/o3/o4-mini\footnote{\url{https://openai.com/index/introducing-o3-and-o4-mini/}}) demonstrate the power of autonomous tool invocation within reasoning chains-of-thought~\cite{liSTARTSelftaughtReasoner2025,jinSearchR1TrainingLLMs2025,wuAgenticReasoningReasoning2025}. 
        A notable advancement in this area is the Deep Video Discovery~\cite{zhangDeepVideoDiscovery2025} agent, which exemplifies this paradigm by leveraging sophisticated Large Language Model (LLM) reasoning to autonomously plan and strategically select search-centric tools over a multi-granular video database, rather than relying on manually designed rigid workflows. 
        This approach offers a compelling reasoning technique that can greatly enhance Infinite Video Understanding, offloading and consolidation. Indeed, the ability to autonomously chain tool calls helps to overcome context and computational limits~\cite{choudhuryZeroShotVideoQuestion2023}, a critical aspect for long-form video understanding, which faces challenges due to extensive temporal-spatial complexity and the limitations of current LLMs in handling dense, hour-long video contexts. 
        In this paradigm, models dynamically use specialized tools on relevant video segments. 
        The Deep Video Discovery agent, for instance, constructs a multi-grained video database and is equipped with a suite of search-centric tools --- Global Browse, Clip Search, and Frame Inspect --- that operate at varying levels of granularity. 
        The outputs from these tools, which include global summaries, clip captions, and original pixel-level frames, build an external, queryable knowledge base, thus avoiding raw data storage limits. 
        Moreover, by basing answers on accurate, visually-grounded evidence collected and iteratively refined through tool usage, rather than relying solely on the latent knowledge of the model, the issue of hallucination could be implicitly mitigated.
    \item \textbf{Novel Benchmarks and Evaluation Methodologies}. 
        Current benchmarks, even for long videos, are based on finite datasets~\cite{zouSecondsHoursReviewing2024,chenLongVILAScalingLongContext2024,chengScalingVideoLanguageModels2025,wuLongViTUInstructionTuning2025,luoVideoRAGVisuallyalignedRetrievalAugmented2024,liuVideoXLProReconstructiveToken2025,chandrasegaranHourVideo1HourVideoLanguage2024} (see Table~\ref{tab:longform_benchmark_summary}). 
        Evaluating Infinite Video Understanding requires new paradigms. 
        Recent contributions to this include the V-RULER benchmark, proposed by VideoRoPE++~\cite{weiVideoRoPEWhatMakes2025,hsiehRULERWhatsReal2024}, which features novel and challenging subtasks such as Needle Retrieval under Distractors (NRD), Multi-Key, Multi-Value (MKMV), Counting, and Ordering, specifically designed to assess fine-grained temporal localization, entity tracking, and robustness in long-context scenarios. 
        The Lengthy Multimodal Stack task further pushes the boundaries for multimodal robustness evaluation. 
        Deep Video Discovery~\cite{zhangDeepVideoDiscovery2025} also comprehensively evaluates on LVBench~\cite{wangLVBenchExtremeLong2024}, LongVideoBench~\cite{wuLongVideoBenchBenchmarkLongcontext2024}, Video-MME~\cite{fuVideoMMEFirstEverComprehensive2024}, and EgoSchema~\cite{mangalamEgoSchemaDiagnosticBenchmark2023}, achieving state-of-the-art performance. 
        Similarly, HoPE~\cite{liHoPEHybridPosition2025} and Video-XL-2~\cite{qinVideoXL2VeryLongVideo2025} rigorously test their models on MLVU~\cite{zhouMLVUBenchmarkingMultitask2025}, LongVideoBench~\cite{wuLongVideoBenchBenchmarkLongcontext2024}, and Video-MME~\cite{fuVideoMMEFirstEverComprehensive2024} across various context lengths, demonstrating advanced capabilities in long video understanding and retrieval. 
        These recent benchmarks serve as crucial steps, but we still need:
        \begin{itemize}
            \item simulated infinite video streams with controllable properties and hidden long-term patterns;
            \item tasks requiring reasoning over past events at arbitrary distances (e.g., ``Based on everything you've seen, why did event $X$ happen?'');
            \item tasks involving predicting future events or trends based on cumulative history;
            \item performance metrics for evaluating the coherence and detail of the model's internal state or memory over time;
            \item performance metrics for evaluating the real-time responsiveness and the ability to process continuous input without falling behind.
        \end{itemize}
        Developing challenging, multi-task benchmarks like extensions of MLVU~\cite{zhouMLVUBenchmarkingMultitask2025} or VNBench-Long~\cite{zhaoNeedleVideoHaystack2025} for infinite depth, but adapted for continuous evaluation, will be critical~\cite{chengScalingVideoLanguageModels2025,wuLongViTUInstructionTuning2025,liVideoChatFlashHierarchicalCompression2025,luoVideoRAGVisuallyalignedRetrievalAugmented2024}.
        Please see more discussion on this topic in \S~\ref{sec:Evaluating}.
\end{itemize}

\section{Beyond Finite Frames: Evaluating Infinite Video Understanding}
\label{sec:Evaluating}

Existing benchmarks for long video understanding predominantly adapt traditional metrics from short-form video and natural language processing (NLP) tasks. 
For example, multiple-choice video question answering (QA) is typically evaluated using accuracy, defined as the proportion of correctly predicted answers. 
Open-ended QA and captioning tasks employ n-gram-based metrics such as BLEU, ROUGE, METEOR, or CIDEr, with recent advancements incorporating large language model (LLM)-based semantic scoring. 
Temporal localization tasks, such as moment retrieval, rely on temporal Intersection-over-Union (IoU) between predicted and ground-truth segments, often paired with recall at various IoU thresholds (e.g., Recall@1 with IoU $\geq 0.5$). 
While these metrics effectively gauge performance for finite-length videos, they falter as video duration and task complexity extend toward unbounded temporal contexts. 
Current benchmarks, designed for constrained video lengths, fail to address the persistent memory and contextual coherence required for infinite-length video streams.

We classify these traditional metrics into three categories: 
\textbf{(1) Classification-based}, 
\textbf{(2) Generation-based}, and 
\textbf{(3) Localization-based}. 
For classification-based tasks, such as multiple-choice QA, accuracy is mathematically expressed as:
\begin{equation}
    \text{Accuracy} = \frac{N_{\text{correct}}}{N},
\end{equation}
where $N_{\text{correct}}$ denotes the number of correct predictions, and $N$ is the total number of questions. 
For generation-based tasks, BLEU serves as a representative metric, calculated as:
\begin{equation}
    \text{BLEU}_N = \exp\left( \sum_{n=1}^{N} \frac{1}{N} \log p_n \right) \cdot \exp\left(1 - \frac{r}{c}\right),
\end{equation}
where $p_n$ represents the precision of $n$-grams, $r$ is the reference length, and $c$ is the candidate length. 
Temporal localization metrics utilize temporal IoU, defined as:
\begin{equation}
    \text{IoU}(t_{\text{pred}}, t_{\text{gt}}) = \frac{|t_{\text{pred}} \cap t_{\text{gt}}|}{|t_{\text{pred}} \cup t_{\text{gt}}|},
\end{equation}
with recall formulated as:
\begin{equation}
    R@K(\tau) = \frac{1}{Q} \sum_{i=1}^{Q} \mathbf{1} \left( \max_{k \leq K} \text{IoU}(p_{ik}, g_i) \geq \tau \right),
\end{equation}
where $Q$ is the number of queries, $p_{ik}$ is the $k$-th predicted segment, $g_i$ is the ground-truth segment, and $\tau$ is the IoU threshold. 
These metrics excel at evaluating pointwise or segment-level performance but fall short in capturing temporal consistency, long-term memory, or evolving context—attributes critical for infinite-length video understanding.

To tackle the distinct challenges of \textit{Infinite Video Understanding}, where models must process and reason over unbounded video streams, we introduce three novel, model-agnostic evaluation metrics:
\begin{itemize}
    \item \textbf{Long-Horizon Consistency Score (LCS)}: 
    This metric evaluates a model's ability to provide consistent responses to identical queries posed at different timestamps across an unbounded video. 
    Given $M$ responses $\{a_1, a_2, \dots, a_M\}$ to the same question, LCS is defined as:
    \begin{equation}
        \text{LCS} = 1 - \frac{1}{M(M-1)} \sum_{i \neq j} \mathbf{1} (a_i \neq a_j),
    \end{equation}
    where $\mathbf{1} (a_i \neq a_j)$ is an indicator function returning 1 if responses $a_i$ and $a_j$ differ, and 0 otherwise. 
    LCS quantifies temporal coherence over extended horizons.

    \item \textbf{Entity Trajectory Recall (ETR)}: 
    This metric assesses a model's capacity to accurately track an entity throughout an infinite-length video. 
    For $K$ annotated entities, ETR is computed as:
    \begin{equation}
        \text{ETR} = \frac{1}{K} \sum_{k=1}^{K} \mathbf{1}(\mathcal{T}_k),
    \end{equation}
    where $\mathcal{T}_k$ indicates successful tracking of the $k$-th entity across the entire video. 
    ETR emphasizes long-term entity recognition and memory retention.

    \item \textbf{Progressive Reasoning Fidelity (PRF)}: 
    This metric measures the logical coherence of a model's intermediate reasoning steps as additional video context accumulates. 
    For a task with $T$ decomposable reasoning steps, PRF is given by:
    \begin{equation}
        \text{PRF} = \frac{1}{T} \sum_{t=1}^{T} \mathbf{1}(\mathcal{R}_t),
    \end{equation}
    where $\mathcal{R}_t$ denotes the event that the $t$-th reasoning step remains valid with increasing context. 
    PRF evaluates a model's ability to dynamically refine reasoning without sacrificing consistency.
\end{itemize}
These metrics --- LCS, ETR, and PRF --- prioritize persistent memory, long-range consistency, and adaptive reasoning fidelity, addressing gaps in traditional metrics. 
By focusing on holistic, temporally integrated understanding, they provide a robust framework for evaluating models in the context of infinite-length video streams.


\section{Research Areas Related to Infinite Video Understanding}

\subsection{Continual Learning / Lifelong Learning / Never-Ending Learning}

The ambition of achieving Infinite Video Understanding represents a fundamental shift in how we approach processing and comprehending visual data streams of arbitrary, potentially unbounded duration. 
As we outline the core challenges and key research directions for Infinite Video Understanding, it is crucial to situate this frontier within the broader landscape of Machine Learning or Deep Learning research, particularly in relation to the established paradigm of Continual Learning, also known as Lifelong Learning or Never-ending Learning~\cite{wangComprehensiveSurveyContinual2024,quRecentAdvancesContinual2024,zhengLifelongLearningLarge2024}. 
While both fields grapple with the complexities of learning from sequential data over extended periods, their specific problem formulations, challenges, and objectives exhibit distinct characteristics. 
Understanding this interplay is vital for leveraging advancements in Continual Learning to accelerate progress towards Infinite Video Understanding.

At its core, Continual Learning focuses on enabling models to sequentially learn new tasks or data distributions without significantly degrading performance on tasks learned previously. 
This addresses the phenomenon of ``catastrophic forgetting'', where training on new data overwrites knowledge acquired from old data~\cite{wangComprehensiveSurveyContinual2024,quRecentAdvancesContinual2024,zhengLifelongLearningLarge2024}. 
The goal is to accumulate knowledge and skills over a lifetime, adapting to new information arriving in sequential order. 
Key challenges in Continual Learning include balancing stability (retaining old knowledge) and plasticity (learning new knowledge), managing memory limitations, and handling diverse task sequences~\cite{wangComprehensiveSurveyContinual2024,quRecentAdvancesContinual2024,zhengLifelongLearningLarge2024}. 
Common techniques employed range from regularization-based methods, knowledge distillation, and memory-based approaches (such as experience replay), to architecture-based solutions~\cite{wangComprehensiveSurveyContinual2024,quRecentAdvancesContinual2024,zhengLifelongLearningLarge2024}.

Infinite Video Understanding shares several fundamental challenges with Continual Learning. 
\begin{itemize}
    \item Both areas must contend with processing data that arrives sequentially and can represent a vast or even infinite volume over time. 
        This necessitates developing architectures capable of efficient processing and managing memory over extended duration, moving beyond traditional batch-based methods. 
    \item Both areas struggle with maintaining coherence and tracking dependencies across temporally distant events or tasks. 
    \item Both areas have the same challenge to handle noisy or variable data, which is particularly prevalent in real-world streams like user-generated video content.
    \item Evaluating performance over such large scales and dynamic conditions also presents significant hurdles for both areas.
\end{itemize}

However, a critical distinction lies in the nature of the input stream and the definition of the learning objective. 
In the classical Continual Learning formulation, learning often occurs in discrete tasks or domains, where the model is explicitly trained on datasets corresponding to distinct knowledge sets. 
The primary objective is typically to perform well on a set of tasks encountered over time, often evaluated by average performance across all tasks seen so far or the ability to distinguish between tasks.  
By contrast, Infinite Video Understanding focuses on a single, continuous video stream. 
\begin{itemize}
    \item While this stream may contain multiple events or segments that could be interpreted as implicit ``tasks'', the fundamental goal is not necessarily to learn a sequence of distinct, separable skills, but rather to build and maintain a persistent, evolving, and coherent understanding of the unfolding video content over arbitrary duration. 
        This involves capabilities like real-time processing, retrospective querying, tracking cumulative information, and predicting future developments based on continuous historical context. 
    \item The video data is dense, multimodal, and temporally highly correlated, unlike the potentially uncorrelated data streams encountered in task-based Continual Learning scenarios. 
    \item Forgetting here means losing temporal coherence, failing to track complex events, or being unable to refer to distant past information within the same continuous stream, rather than the complete loss of a previously mastered distinct task.
\end{itemize}

Despite these distinctions, the research methodologies and techniques developed in Continual Learning offer significant potential to help advancing Infinite Video Understanding. 
Concepts from Continual Learning, such as memory-based approaches for storing and replaying past data or knowledge~\cite{wangComprehensiveSurveyContinual2024,quRecentAdvancesContinual2024,zhengLifelongLearningLarge2024}, are directly relevant to the need for persistent memory and knowledge consolidation in Infinite Video Understanding architectures. 
Techniques for managing the balance between stability and plasticity~\cite{wangComprehensiveSurveyContinual2024,quRecentAdvancesContinual2024,zhengLifelongLearningLarge2024} can inform how Infinite Video Understanding models continuously update their understanding while retaining the context of past events. 
Architecture-based Continual Learning methods that allow for dynamic expansion or adaptation~\cite{wangComprehensiveSurveyContinual2024,quRecentAdvancesContinual2024,zhengLifelongLearningLarge2024} could provide blueprints for building models capable of processing and understanding continuous video streams without a predefined endpoint. 
The application of Continual Learning to video-specific tasks like temporal action segmentation or video classification, and recent work on Continual Learning for LLMs or MLLMs~\cite{wangComprehensiveSurveyContinual2024,quRecentAdvancesContinual2024,zhengLifelongLearningLarge2024}, highlight explicit connections and potential transfer pathways for techniques and insights. 
Retrieval-Augmented Generation (RAG), already explored in long video understanding using auxiliary texts~\cite{luoVideoRAGVisuallyalignedRetrievalAugmented2024}, aligns conceptually with external knowledge approaches in Continual Learning~\cite{wangComprehensiveSurveyContinual2024,quRecentAdvancesContinual2024,zhengLifelongLearningLarge2024}.

In essence, Infinite Video Understanding can be viewed as a demanding and complex instance of the broader Lifelong, Continual Learning problem, specialized for the unique characteristics of continuous, high-density video streams and the challenging objective of maintaining a deep, evolving comprehension over arbitrary timescales. 
While Continual Learning provides a rich foundation of techniques and theoretical understanding regarding learning from sequential data and mitigating forgetting, Infinite Video Understanding introduces orthogonal complexities related to spatial-temporal coherence, multi-modality fusion in a streaming setting, and the sheer scale and continuity of the data source. 
Undoubtedly, progress in one area will inform and accelerate breakthroughs in the other, pushing the boundaries of intelligent systems capable of operating effectively in dynamic, data-rich environments.

\subsection{Universal Personal Assistants}
\label{sec:Universal_Personal_Assistants}

The research towards Infinite Video Understanding may find a compelling real-world impetus in the rapidly evolving landscape of egocentric Universal Personal Assistants\footnote{\url{https://blog.google/technology/google-deepmind/gemini-universal-ai-assistant/}}. 
Projects such as EgoLife~\cite{yangEgoLifeEgocentricLife2025} enabled by wearable smart devices like Meta's AI glasses from its Project Aria~\cite{engelProjectAriaNew2023} exemplify the trajectory towards pervasive, wearable AI systems. 
These systems, often embodied in AI glasses, capture continuous, multimodal data streams from the wearer's first-person perspective over extended periods --- minutes, hours, days, and potentially a lifetime. 
This continuous, first-person perspective video constitutes precisely the kind of arbitrary-duration stream that Infinite Video Understanding strives to comprehend. 
Indeed, the concept of understanding the entirety of one's life experience, from birth to death, can actually be framed as the ultimate first-person, Infinite Video Understanding problem.

A Universal Personal Assistant operating via AI glasses fundamentally relies on the ability to maintain a coherent, evolving understanding of the user's ongoing experience as captured through egocentric sensors like cameras. 
This necessitates capabilities that extend far beyond the finite video understanding tasks typical of current state-of-the-art models, which encounter significant constraints when faced with the sheer volume of visual tokens from lengthy sequences. 
Systems like EgoLife's EgoButler~\cite{yangEgoLifeEgocentricLife2025} are beginning to address this challenge for ``ultra-long'' contexts spanning a week, employing approaches like dense multimodal captioning and Retrieval-Augmented Generation (RAG)~\cite{luoVideoRAGVisuallyalignedRetrievalAugmented2024} over a multi-level memory bank to support long-context question answering. 
Meta's Project Aria~\cite{engelProjectAriaNew2023} provides a hardware and software foundation specifically designed to foster research into context-aware, personalized AI based on egocentric multimodal data, explicitly identifying summarization and question answering over ``longer time periods'' as crucial components. 
Current methods, even on benchmarks like HourVideo~\cite{chandrasegaranHourVideo1HourVideoLanguage2024} which focuses on 1-hour egocentric video understanding, exhibit performance significantly below human level.

The technical hurdles encountered in developing such egocentric Universal Personal Assistants directly mirror the core challenges we posit for achieving Infinite Video Understanding. 
For instance, the requirement for ``long-context question answering over extensive temporal information'' in EgoLifeQA~\cite{yangEgoLifeEgocentricLife2025} underscores the fundamental Infinite Video Understanding challenge of maintaining temporal coherence, tracking complex events, and preserving fine-grained details over extended periods. 
The EgoButler system's reliance on EgoRAG~\cite{yangEgoLifeEgocentricLife2025} to retrieve relevant evidence across video segments to mitigate issues faced by models attempting to process ultra-long videos in chunks highlights the need for persistent memory and knowledge consolidation architectures central to Infinite Video Understanding research directions. 
Managing the continuous inflow of high-density egocentric data and overcoming memory and computational constraints aligns with Infinite Video Understanding's need for efficient streaming, incremental processing, and hierarchical or adaptive spatiotemporal representations. 
Moreover, the challenge of handling noisy and misaligned egocentric data is a critical aspect of Infinite Video Understanding's requirement for robustness against input variability and concept drift. 
Evaluating the understanding maintained by these systems over potentially week-long periods, as attempted with EgoLifeQA, points directly to the need for novel benchmarks and evaluation methodologies capable of assessing the ability of video comprehension at unprecedented temporal scales~\cite{yangEgoLifeEgocentricLife2025,engelProjectAriaNew2023} and examining the associated ethical issues~\cite{gabrielEthicsAdvancedAI2024}. 

It is important to note that while Universal Personal Assistants represent a powerful application domain and key motivation for Infinite Video Understanding, they are not synonymous. 
Infinite Video Understanding is a foundational video understanding capability --- the ability to process and comprehend any video stream of arbitrary length. 
Universal Personal Assistants are specific systems that leverage such a capability to perform user-oriented tasks based on a particular type of video stream (first-person egocentric data). 
The development of robust, reliable, and truly helpful Universal Personal Assistants will necessitate fundamental advances in Infinite Video Understanding, pushing the boundaries of long-context understanding from current ``ultra-long'' achievements towards the continuous processing and comprehension of unbounded visual histories. 
Therefore, the pursuit of Infinite Video Understanding serves as a vital research \emph{north star}, driven in no small part by the compelling future vision of intelligent systems embedded within our daily lives via form factors such as AI glasses made by Meta\footnote{\url{https://www.projectaria.com/}}, Google\footnote{\url{https://deepmind.google/models/project-astra/}}, and so on.

\subsection{Infinite Games}

In advancing the discourse on Infinite Video Understanding, it is instructive to draw a compelling conceptual parallel with the philosophical framework presented by James P. Carse in his seminal work, \emph{Finite and Infinite Games}\footnote{\url{https://www.canva.com/design/DAGp0iRLk9g/8QLkIDov8ez1q6VvO8nnpQ/view}}~\cite{carseFiniteInfiniteGames2011}. 
Carse posits that there exist two fundamental modalities of engagement: ``finite games'' and ``infinite games''. 
A finite game is defined by its objective: to win. 
It is played within pre-established rules and boundaries, culminating in a definitive victor and vanquished. 
In stark contrast, an infinite game is played for the sole purpose of continuing the play. 
Its participants are not bound by rigid rules or fixed endpoints; rather, they may continually adapt the rules and boundaries to ensure the game's perpetuation. 
This distinction, Carse argues, extends across diverse spheres of human endeavour, distinguishing between ``theatrical'' roles performed according to a script and ``dramatic'' choices enacted in the present.

Applying this profound distinction to the domain of video understanding illuminates the fundamental paradigm shift necessitated by Infinite Video Understanding.
\begin{itemize}
  \item \textbf{Current Long Video Understanding as a ``Finite Game''}.
    The prevailing research in long video understanding, despite its advancements, largely operates within the confines of a finite game paradigm. 
    Models are primarily developed and evaluated on benchmarks of limited, albeit extended, duration, typically measured in minutes or a few hours. 
    The objective is often to achieve a maximal score on a specific task (e.g., captioning, question-answering, event detection) for a predefined, bounded video segment. 
    Current architectures, including MLLMs, are inherently constrained by finite context windows and suffer from significant computational and memory limitations when faced with the sheer volume of visual tokens from lengthy sequences. These limitations represent the explicit boundaries and rules within which the ``game'' of long video understanding is presently played. 
    Success is defined by ``winning'' on these pre-set benchmarks, analogous to a finite game yielding a winner. 
    Furthermore, processing often follows a batch-oriented paradigm, where a finite input is processed from beginning to end to yield a result, a characteristic aligned with the fixed script of a ``theatrical'' performance.
  \item \textbf{Proposed Infinite Video Understanding as an ``Infinite Game''}.
    The vision for Infinite Video Understanding fundamentally transcends this finite game perspective, embodying the principles of an infinite game. 
    The core objective of Infinite Video Understanding is not to achieve a singular winning state on a bounded video segment, but rather to continuously process, understand, and reason about video streams of arbitrary, potentially never-ending duration. 
    This aligns precisely with the infinite game's purpose of ``continuing the play'' indefinitely. 
    Infinite Video Understanding necessitates a conceptual departure from merely scaling up existing techniques: it is a blue-sky research objective that defines a \emph{north star} for the multimedia community, urging innovation beyond incremental improvements.

    The inherent requirements of Infinite Video Understanding reflect the dynamic nature of an infinite game.
    \begin{itemize}
      \item \textbf{Continuous Evolution and Adaptation}. 
      Unlike finite games with fixed rules, Infinite Video Understanding models must be designed to continuously update their knowledge and reasoning capabilities. 
      This includes the imperative to adapt to changing patterns or concepts within the stream over indefinite periods and to exhibit resilience to noise and concept drift. 
      This mirrors the infinite game's capacity to change rules and play with boundaries to sustain its existence.
      \item \textbf{Persistent Understanding, Not Terminal Output}. 
      The goal shifts from producing a final summary or answer for a discrete video to maintaining a persistent, evolving understanding of the continuous video data stream. 
      This entails processing data incrementally, building abstract long-term memories, and selectively retrieving past information without re-processing entire histories. 
      Such a focus on the process of ongoing comprehension, rather than a definitive, terminal ``win'', is a hallmark of the infinite game.
      \item \textbf{Dynamic and Dramatic Engagement}. 
      Infinite Video Understanding moves beyond the ``theatrical'' performance on pre-defined datasets to a ``dramatic'' engagement with the unbounded world, where models continuously enact and update their understanding based on incoming, unforeseen data. 
      It represents an open problem that demands novel and exploratory solutions, inviting a fundamental re-thinking of existing paradigms.
    \end{itemize}
\end{itemize}

In essence, framing Infinite Video Understanding through the lens of Carse's infinite game provides a potent conceptual framework. 
It underscores that our ultimate aspiration is not merely to process longer videos more efficiently, but to cultivate intelligent systems capable of sustained, adaptive, and perpetual understanding of the video streams that increasingly mediate our interaction with the world. 
This is a grand challenge worthy of concerted pursuit, promising transformative impact.

\section{Connections to AGI}

As we in the multimedia research community navigate the evolving landscape of AI, particularly moving into what has been termed the ``second half''\footnote{\url{https://ysymyth.github.io/The-Second-Half/}} or the ``era of experience''~\cite{silverWelcomeEraExperience}, the focus is shifting from merely developing novel training methods to defining challenging, real-world problems and evaluation setups that push beyond existing benchmarks. 
This new era necessitates agents that learn from their own interaction with the environment through continuous streams of experience, rather than solely from static, human-curated datasets. 
Within this context, Infinite Video Understanding emerges as a profoundly relevant and arguably essential research frontier on the path towards achieving Artificial General Intelligence (AGI).

The pursuit of Infinite Video Understanding is deeply intertwined with the grand aspiration of AGI in several key aspects.
\begin{itemize}
    \item A fundamental characteristic of agents in the era of experience is their ability to inhabit streams of experience over extended periods. 
        Infinite Video Understanding directly addresses the challenge of perceiving and comprehending visual and multimodal information from such streams of arbitrary, potentially unbounded duration. 
        Real-world applications like AI glasses (see \S~\ref{sec:Universal_Personal_Assistants}), automated driving, robot perception, and surveillance all involve processing continuous video feeds, which are core components of an agent's experiential stream.
    \item Achieving superhuman capabilities often lies beyond the limits of knowledge extracted from existing human data and static procedures. 
        Infinite Video Understanding, by definition, requires models that can operate continuously and maintain a persistent, evolving understanding without relying on finite context windows or reprocessing historical data from scratch. 
        This necessitates fundamentally new architectures for memory consolidation, storage, and retrieval over indefinite time scales, capabilities that are also crucial for AGI agents operating over a lifetime of experience. 
        Concepts like continuous consolidation of short-term observations into long-term memory (LTM) and Retrieval-Augmented Generation (RAG) are explored in both Infinite Video Understanding and the broader context of experiential learning.
    \item Agents in the era of experience are expected to interact with the real world, with actions and observations grounded in their environment. 
        Video is a primary modality for grounding perception in dynamic, real-world environments, thus Infinite Video Understanding is essential for these agents to build a coherent understanding of the visual world they inhabit, reason about complex events unfolding over long periods, and ultimately ground their internal reasoning processes in reality rather than solely imitating human thought. 
        While current reasoning often mimics human's chains-of-thought, progress towards AGI will likely require agents to discover more efficient, non-human reasoning mechanisms grounded in real-world interaction. 
        Infinite Video Understanding provides a critical test-bed for developing such grounded perceptual reasoning capabilities.
    \item The ``utility problem'', where AI excels on benchmarks but struggles with real-world usefulness, is attributed partly to evaluation setups differing from reality. 
        The second half of AI research calls for developing novel evaluation setups and tasks for real-world utility. 
        Infinite Video Understanding, with its focus on processing continuous, noisy, and potentially misaligned real-world video streams, is inherently aligned with real-world demands. 
        Developing effective benchmarks for Infinite Video Understanding that capture continuous understanding and reasoning over vast temporal horizons presents a significant challenge that directly contributes to addressing the utility problem for intelligent agents.
    \item Progress in AI has historically been driven by unlocking and effectively utilizing new sources of data\footnote{\url{https://blog.jxmo.io/p/there-are-no-new-ideas-in-ai-only}}\textsuperscript{,}\footnote{\url{https://kevinlu.ai/the-only-important-technology-is-the-internet}}. 
        Video, particularly egocentric video from wearable smart devices like AI glasses, represents a massive, under-exploited data stream that far exceeds existing text corpora. 
        Developing Infinite Video Understanding capabilities is necessary to harness this data at scale, potentially providing the critical data-driven boost needed for the next paradigm shift towards more capable, perhaps superhuman, intelligence. 
\end{itemize}

While deeply connected, it is important to distinguish Infinite Video Understanding as a specific problem domain from AGI as a broad goal.
\begin{itemize}
    \item AGI aims for general intellectual capability across a wide range of tasks and domains, rivaling or surpassing human intelligence. 
        Infinite Video Understanding, while complex and multimodal, is fundamentally focused on the specific task of understanding video streams. 
        It is a critical component for AGI systems operating in dynamic, visually-rich environments, but it does not encompass the full spectrum of cognitive abilities associated with AGI (e.g., abstract mathematical reasoning, complex social interaction beyond interpreting visual cues, planning in abstract spaces not directly tied to perceptual streams, and so on).
    \item Infinite Video Understanding places particular emphasis on the unique challenges posed by continuous, high-density video data, such as managing immense data volume, maintaining temporal coherence over arbitrary lengths, and developing streaming or incremental processing architectures. 
        While AGI also requires handling diverse data and temporal dynamics, the core problems addressed by Infinite Video Understanding are specific to continuous sensory input streams.
    \item Solving Infinite Video Understanding would represent a monumental step forward in AGI, enabling capabilities previously impossible. 
        However, a system with perfect Infinite Video Understanding might still lack other critical components of AGI, such as sophisticated motor control for robots, advanced natural language generation and dialogue management, or the ability to perform complex reasoning across disparate knowledge domains outside of what is observed in the video stream.
\end{itemize}

In essence, Infinite Video Understanding represents a fundamental and currently unsolved open problem within the perceptual domain that is a necessary, albeit not solely sufficient, condition for achieving AGI, particularly for intelligent agents designed to operate autonomously and learn from experience in the real world. 
By focusing research efforts on this blue-sky challenge, we drive innovation in critical areas like continuous processing, persistent memory, and grounded reasoning, laying essential groundwork for the development of truly capable and generally intelligent systems.

\section{Conclusion}

Infinite Video Understanding is a vision that describes a significant, currently open problem in multimedia research --- how to build systems that truly understand video over unbounded time. 
It requires novel, exploratory solutions that fundamentally rethink existing paradigms. 
The innovative architectural designs in Video-XL-2~\cite{qinVideoXL2VeryLongVideo2025}, the principled positional encoding advancements in HoPE~\cite{liHoPEHybridPosition2025} and VideoRoPE++~\cite{weiVideoRoPEWhatMakes2025}, and the autonomous agentic search framework of Deep Video Discovery~\cite{zhangDeepVideoDiscovery2025} all exemplify the directions that our community has started pursuing to embark on this journey. 
While initial work might not present empirical results outperforming state-of-the-art on existing finite benchmarks (though it could lay the groundwork), the focus is on the scientific argument, the identification of the problem, and the proposed research trajectory. 
It invites connecting classical concepts (like memory systems or hierarchical representations) to current developments in MLLMs to forge entirely new directions. 
Infinite Video Understanding is a challenge that, if addressed, promises fundamental, significant advances in multimedia research and its applications.

In recent years, the field of multimedia research has made tremendous strides in video understanding, particularly with the advent of MLLMs. 
However, the challenge of processing and comprehending videos of significant length remains a critical barrier to unlocking the full potential of visual AI in real-world applications. 
We argue that setting our sights on Infinite Video Understanding as a long-term research goal provides the necessary ambition and direction to drive the next generation of breakthroughs. 
Infinite Video Understanding demands a departure from simply scaling up existing models and calls for fundamental innovation in areas spanning system architecture, memory management, data representation, processing paradigms, and evaluation methodologies. 
It represents a complex, multi-faceted open problem that necessitates a concentrated effort from the multimedia research community. 
By embracing Infinite Video Understanding as our \emph{north star}, we can push the boundaries of what is possible, enabling intelligent systems that can truly perceive, understand, and interact with a world increasingly mediated by continuous video streams. 
We believe that this is a grand challenge worthy of pursuit, promising transformative impact across numerous domains.

\bibliographystyle{ACM-Reference-Format}
\bibliography{ref_video-understanding,ref_continual-learning}

\end{document}